\documentclass[5p,twocolumn]{elsarticle}

\usepackage{graphicx}
\usepackage{picinpar}
\usepackage{picins}
\usepackage{amsmath}
\usepackage{url}
\usepackage{flushend}
\usepackage[utf8]{inputenc}
\usepackage{colortbl}
\usepackage{soul}
\usepackage{multirow}
\usepackage{pifont}
\usepackage{color}
\usepackage{alltt}
\usepackage[hidelinks]{hyperref}
\usepackage{enumerate}
\usepackage{siunitx}
\usepackage{breakurl}
\usepackage{epstopdf}
\usepackage{pbox}
\usepackage{subfigure}
\usepackage{amsfonts}
\usepackage{algorithm}
\usepackage{algorithmic}
\usepackage{setspace}
 
\usepackage{lineno,hyperref}

\journal{Neuro Computing}

\begin{document}

\begin{frontmatter}

\title{Training Deep Neural Networks for Wireless Sensor Networks Using Loosely and Weakly Labeled Images}



\author[mymainaddress,mysecondaryaddress]{Qianwei Zhou}
\ead{zhouqianweischolar@gmail.com}

\author[mymainaddress,mysecondaryaddress]{Yuhang Chen}
\ead{xiaobai456123@gmail.com}

\author[mythridaddress]{Baoqing Li}
\ead{ libq@mail.sim.ac.cn}

\author[mymainaddress,mysecondaryaddress]{ Xiaoxin Li}
\ead{mordekai@zjut.edu.cn}
 
\author[mymainaddress,mysecondaryaddress]{ Chen Zhou}
\ead{hcqylym@gmail.com}

\author[mythridaddress]{Jingchang Huang}
\ead{ jchhuang@mail.ustc.edu.cn}

\author[mymainaddress,mysecondaryaddress]{ Haigen Hu \corref{cor1}}
\ead{hghuzj@gmail.com}
 
\address[mymainaddress]{College of Computer Science and Technology, Zhejiang University of Technology, Hangzhou 310023, China}
\address[mysecondaryaddress]{Key Laboratory of Visual Media Intelligent Processing Technology of Zhejiang Province, Hangzhou 310023, China}
\address[mythridaddress]{Shanghai Institute of Microsystem and Information Technology, Chinese Academy of Sciences, Shanghai 200050, China}

\cortext[cor1]{Corresponding Author}
\fntext[myfootnote]{Accepted by Neuro Computing on September 20, 2020. © 2020. This manuscript version is made available under the CC-BY-NC-ND 4.0 license http://creativecommons.org/licenses/by-nc-nd/4.0/}

\begin{abstract}
Although deep learning has achieved remarkable successes over the past years, few reports have been published about applying deep neural networks to Wireless Sensor Networks (WSNs) for image targets recognition where data, energy, computation resources are limited. In this work, a Cost-Effective Domain Generalization (CEDG) algorithm has been proposed to train an efficient network with minimum labor requirements. CEDG transfers networks from a publicly available source domain to an application-specific target domain through an automatically allocated synthetic domain. The target domain is isolated from parameters tuning and used for model selection and testing only. The target domain is significantly different from the source domain because it has new target categories and is consisted of low-quality images that are out of focus, low in resolution, low in illumination, low in photographing angle. The trained network has about 7M (ResNet-20 is about 41M) multiplications per prediction that is small enough to allow a digital signal processor chip to do real-time recognitions in our WSN. The category-level averaged error on the unseen and unbalanced target domain has been decreased by 41.12\%.
\end{abstract}

\begin{keyword}
Deep Neural Networks \sep Wireless Sensor Networks  \sep Automated Data Labeling \sep  Image Recognition \sep  Transfer Learning \sep Model Compression.
\end{keyword}

\end{frontmatter}

\section{Introduction}\label{sec:introduction}
\begin{figure*}[htb]
\centering
  \includegraphics[width=1\textwidth]{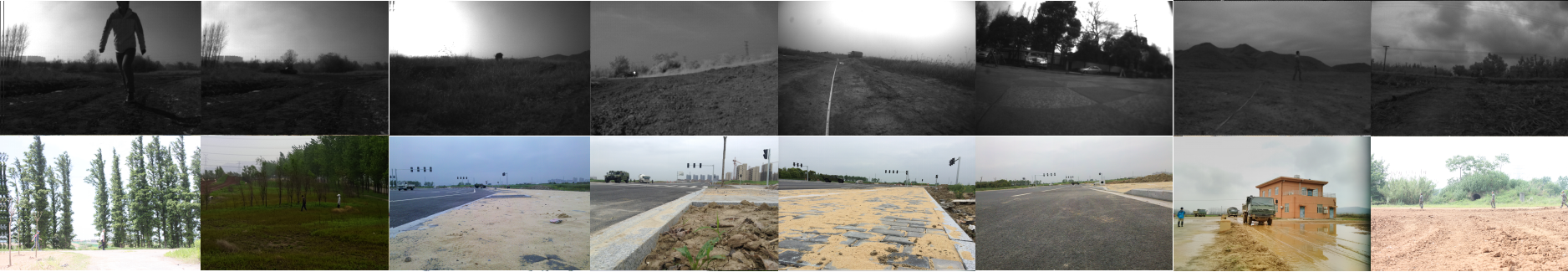} %
  \caption{ Image samples of mixed objects under different illumination. The first row shows low-lighting grey images.} 
  \label{fig1}
\end{figure*}

Wireless sensor networks (WSNs) typically are designed to detect and identify neighboring objects in wild~\cite{ws3,ws8,ws5,ws9} with sound or vibration sensors in the form of single~\cite{ws2} or microarrays~\cite{ws6}. The sound or vibration sensor has many advantages~\cite{ws4,ws1,ws2}, such as low cost, low energy consumption, and relatively low in algorithm complexity. However, they are unsuitable for mixed objects detection because their spatial resolutions are usually too low to distinguish each person in a group of pedestrians. To overcome this shortage, we have employed cameras in our WSNs which has been proved to be effective for dense targets identification~\cite{ws10}. Unfortunately, images captured by WSNs are noisy, such as low in illumination, resolution and photographing angle, which are different from most publicly available datasets. Because the severe limitation in data and resources, despite the rapid development in deep learning~\cite{zeng2019improved,zeng2018facial,zeng2020deep,zhou2020residual,hu2019mc,hu2019background,zhou2019towards,hu2018fast,kong2019classification,an2019generalization,li2019constrained}, WSN-applicable deep-learning-based image classification algorithms evolve slowly. So, cost-effective dataset construction methods are needed urgently to build datasets that corresponding to specific WSN applications. Several random images of the target application (target domain) that was captured during our field experiments have been shown in Fig.~\ref{fig1}, where targets like persons and cars are hard to identify. 

\par
Because of limited communication bandwidth, WSNs cannot run deep neural networks (DNNs) in a remote cloud (or fog) which is a common strategy for embedded devices~\cite{tii1,tii2,tii3,tmc1,tmc2}. To run DNNs in such devices locally~\cite{nc1,nc2}, a training strategy is wanted to cut computation costs without decreasing identification accuracy significantly. Fortunately, Han et al.~\cite{comp1,comp2} have pointed out that only parts of weight parameters in neural networks are playing essential roles during predictions. Therefore, it is possible to train an efficient DNN for WSNs with fewer parameters if we can fully utilize key weight parameters. 

\par
Because fewer parameters mean higher utilization of key parameters and usually lead to less computational costs, we investigated model compression approaches that can be separated into four categories~\cite{ref2}: 1) Parameter Pruning and Sharing~\cite{ref4}, 2) Low-Rank Factorization~\cite{ref32}, 3) Transferred Convolutional Filters~\cite{ref12}, 4) Knowledge Distillation (KD)~\cite{ref14,md}. The compression rates of first three approaches are limited. Some of them focus on reducing storage requirements that do not help in reducing the number of multiplications. The fourth method can train a network with much less parameters which can be further compressed by other three methods if necessary. 

\par
Knowledge Distillation (KD) was first proposed by Bucilu et al.~\cite{md}, which is able to train fast, compact network to mimic better performing but slow and complex network. Usually, the compact network is trained to mimic the high-level features of the complex network. Because first, high level features are highly abstracted that should not be changed during model compression; otherwise, small differences between the features of the compact network and the complex network can lead to big differences in image classification. Second, high level features of image-classification networks are usually low in dimension; with a low dimensional output, one can build slim and compact networks easily. The model compression rate can be very big since the compact network has different structure from the complex network. For example, FitNets~\cite{ref15} trained a network whose parameter number is one-36th of a complex state-of-the-art network. 

\par
Data plays an increasingly important role in neural network training. Sun et al.~\cite{ref31} found that model performance on vision tasks increases logarithmically with training data size. However, building a clean, impartial, diverse dataset is a tremendous challenge. In recent years, studies have been done to apply weakly labeled techniques to construct datasets. Xu et al.~\cite{ref26} referenced weakly labeled technology to create a clean face dataset, which based on the continuity structure of face images of same identity. With the newly constructed face dataset, face recognition accuracy was improved massively. Han et al.~\cite{ref27} proposed a method which based on generative adversarial networks to produce images that are high in quality. The main drawback of traditional methods is that images used to construct the new datasets still require a lot of manpower to label. But in this paper, loosely and weakly labeled images are used to construct new datasets which are downloaded from internet and do not require additional manpower to label. 

\par
In this paper, we propose Cost Effective Domain Generalization (CEDG) algorithm that yields low computational complexity deep neural networks with minimum requirements on datasets. The network has parameters less than 164K for feature extraction and requires about 7 million multiplications per prediction (ResNet-20\footnote{https://github.com/gcr/torch-residual-networks} requires 41 million.). The trained network has reached 87.20\% test accuracy on a common domain (CIFAR-10~\cite{cifar10}), and 87.07\% category-level averaged global generalization accuracy on an unseen target domain. The problem is hard because the target domain: 1) is poor in picture quality; 2) has different image categories from the common domain; 3) cannot be used as train set for the limited image samples.

\par
Our main contribution is that we proposed Cost-Effective Domain Generalization (CEDG) method to solve the challenges of images classification in WSNs (e.g. limited data and computation resources). Detailed contributions are as follows. 
\begin{enumerate}[1)]
\item We have concluded an efficient procedure to train a small-scale but deep network. Comparing to ResNet-20, the network is 40\% smaller (164K parameters vs. 273K), 83\% faster (7M multiplications vs. 41M) and 142\% deeper (46 convolutional layers vs. 19).
\item We have developed a labor-saving method to build a specific synthetic domain quickly from loosely and weakly labeled images. By loosely, we mean labels of images are not 100\% correct. By weakly, we mean the exact locations of the interested targets in the images are unknown.
\item We have specially designed data augmentation methods that can transfer the synthetic domain to the target domain so as the classifier can generalize well on the target domain. 
\end{enumerate}

\par
This paper is organized as follows.  
The theory of CEDG algorithm has been analyzed in Section~\ref{secCEDG}. In Section~\ref{secIMP}, implementations of CEDG have been presented and the experiment results have been shown in Section~\ref{secEXP} followed by the conclusions in Section~\ref{secCONCLU}.

\section{Cost-Effective Domain Generalization}\label{secCEDG}
In this section, we have presented the basic definition and detailed analysis of the proposed Cost-Effective Domain Generalization (CEDG) algorithm. Please see Section~\ref{secIMP} for implementation details.

\begin{figure*}[htb]
\centering
\subfigure[Stage 1: Representation Distillation]{  
\includegraphics[width=3.68cm]{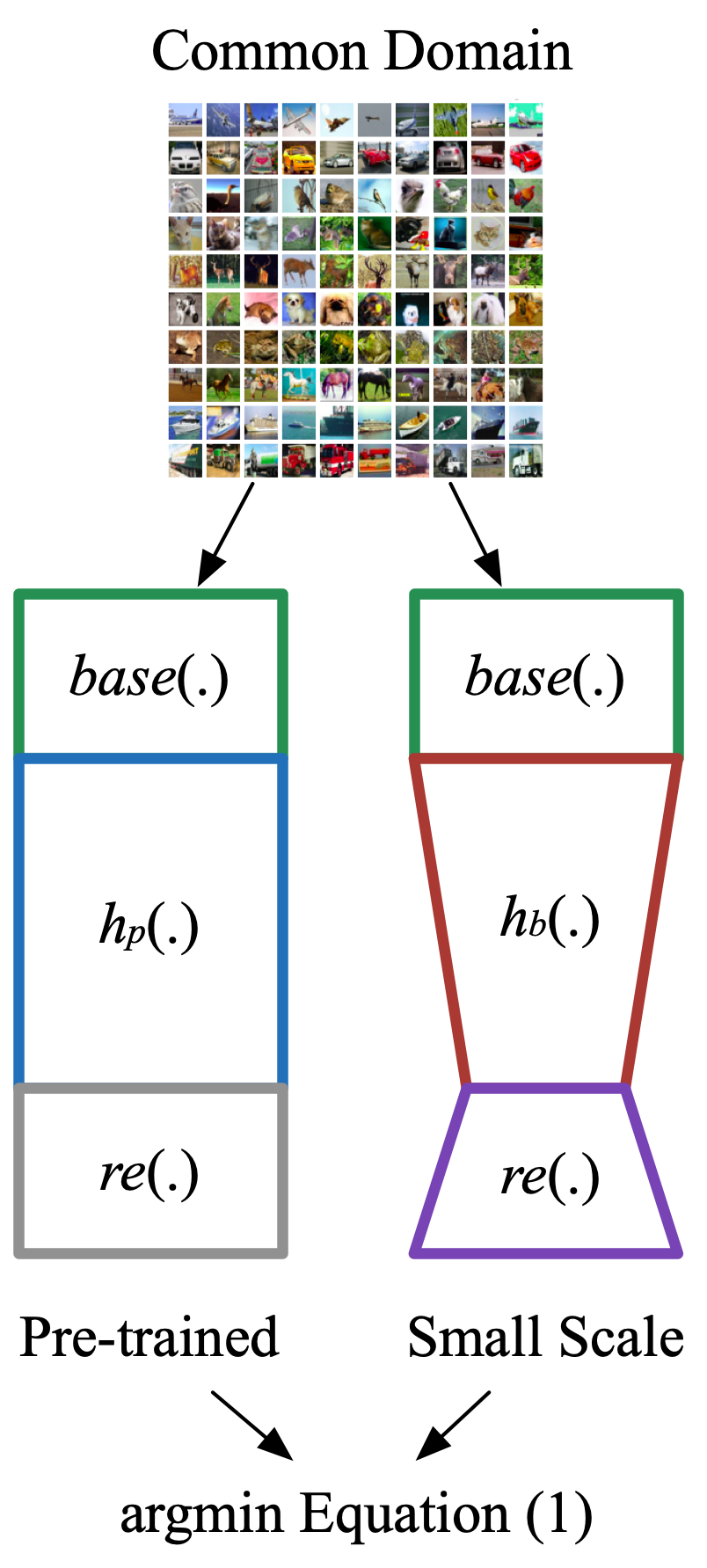}}
\hspace{0.1cm}    
\subfigure[Stage 2: Synthetic Domain $D_s$]{
\includegraphics[width=5.65cm]{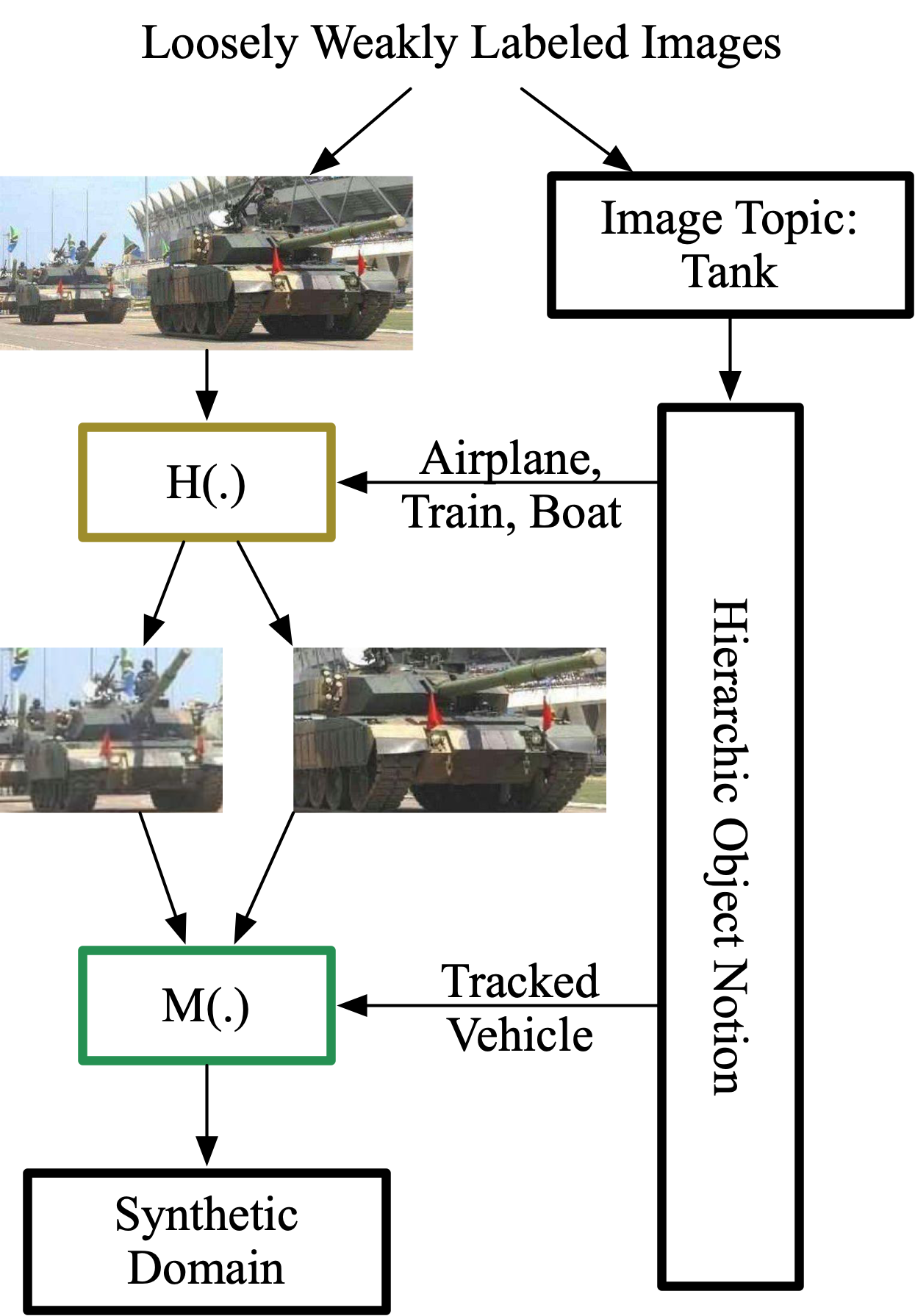}}
\hspace{0.1cm}
\subfigure[Stage 3: Classifier Training]{
\includegraphics[width=3.85cm]{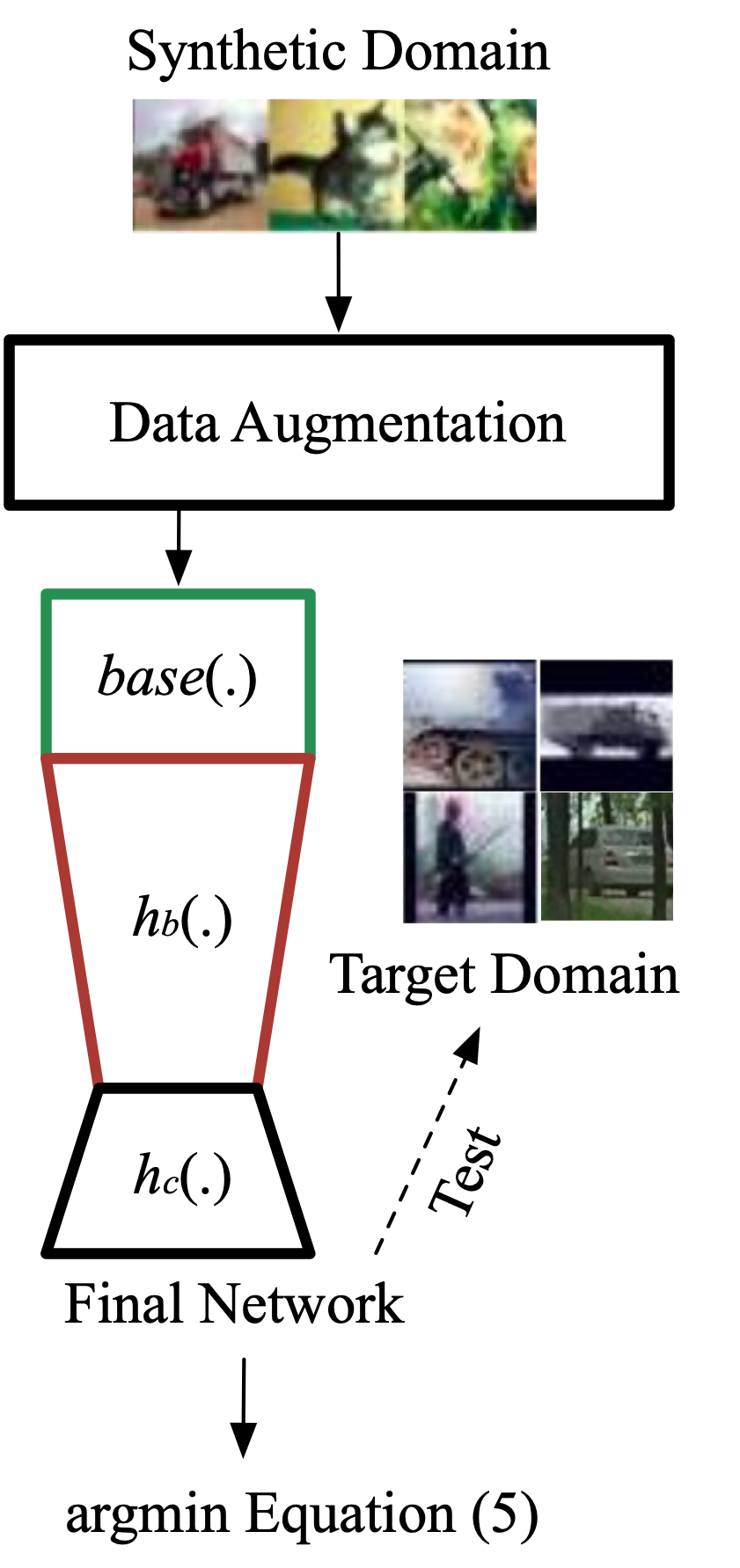}}
\hspace{0.1cm}
\caption{The diagram of CEDG. (a) An efficient network is guided by a pre-trained network on Common Domain. (b) The collected loosely and weakly labeled internet images are converted into images of Synthetic Domain $D_s$ by using Hierarchic Object Notion and Instance Segmentation tool ($\mathbb{H}(.)$) that was designed for other none-related domains. (c) By using the specially designed data augmentation methods, the Final Network is trained with images of Synthetic Domain $D_s$ and is tested on Target Domain which is very different from Common Domain and Synthetic Domain.}
\label{fig2}
\end{figure*}

\subsection{Problem Definition}
The wild images classification problem can be considered as a kind of domain representation learning. Let $X$ and $Y$ be the input feature and label spaces respectively. A domain defined on $X$ $\times$ $Y$ can be described with a joint probability distribution $\mathbb{P}(X,Y)$. For simplicity, we denote the joint  probability distribution $\mathbb{P}^j(X^j,Y^j)$ of a domain j as $\mathbb{P}^j$. The distribution $\mathbb{P}^j$ is associated with a domain $D_j = \{x_i^j,y_i^j\}_{i=1}^{n^j}$ of $n^j$ labeled examples, where $(x_i^j,y_i^j) \sim \mathbb{P}^j$ and $y_i^j \in \mathcal{C}_j$, $\mathcal{C}_j= \{1,2,\ldots,c_j \}$. A target domain can be $D_t = \{x_i^t,y_i^t\}_{i=1}^{n^t}$ , where domain size equal to $n^t$ and $y_i^t \in \mathcal{C}_t$, $\mathcal{C}_t= \{1,2,\ldots,c_t \}$. 

\par 
In this work, a common domain, a synthetic domain, and a target domain have been employed to deal with the image targets classification challenge in WSNs. The target domain $D_t$ is a set of real-world images that are captured from field experiments and used to evaluate the performance of learned representations. The common domain, namely $D_c$, contains a large set of images and labels but its target categories are different from $D_t$. We use publicly available image dataset as $D_c$ to train a computational efficient feature extraction network. The synthetic domain $D_s$ contains lots of loosely labeled images, which is built by a labor-saving automatic algorithm with same target categories as $D_t$. The goal of this work is to learn a classification function $f : X \to Y$ in the synthetic domain $D_s$ and then test the classification function in the target domain $D_t$. The classification function will work well in $D_t$ because the distribution gap between $D_s$ and $D_t$ can be closed by specially designed preprocessing methods.

\subsection{Proposed Method}
We assume there is a common representation or relationship on the synthetic domain with a distribution $\mathbb{P}^s(h(X^s)|Y^s)\mathbb{P}^{s}(Y^s)$, namely $\mathbb{P}^s_h$, that is similar to the target domain, $\mathbb{P}^t(h(X^t)|Y^t)\mathbb{P}^{t}(Y^t)$, or $\mathbb{P}^t_h$ for short. $h(.)$ is a network that can extract the representation. Our goal is to approximate $\mathbb{P}^t_h$ with $\mathbb{P}^s_h$ by training the representation extractor $h(.)$ where $\mathbb{P}^s(Y^s)$ is similar to $\mathbb{P}^{t}(Y^t)$. To obtain computational efficient $h(.)$, we have developed a 3-stage algorithm, namely Cost-Effective Domain Generalization (CEDG). 

\subsubsection{Stage 1: Representation Distillation} 
The first stage is to learn an efficient network $h_b(.)$ from a large pre-trained network in the common domain $D_c$ by imposing Knowledge Distillation (KD)\cite{ref15} constraint as shown in equation~\eqref{eq1}.
\begin{align}
L_{rp} =\frac{1}{2} \|h_p(base(x))-h_b(base(x)))\|_2^2 \label{eq1}
\end{align}
Where $base(x)$ is the bottom layers of the pre-trained network with fixed weight, $h_p(base(x))$ represents the feature extracted by the pre-trained network which usually is the last convolutional layer of the pre-trained network, $h_b(.)$ has adjustable parameters. High-level features (the output of $h_p(.)$) usually are low in dimension. So, $h_b(.)$ can be slim and compact. Plus, high-level or low-dimensional features are the abstract of the input image which means that small differences between the outputs of $h_p(base(x))$ and $h_b(base(x))$ can lead to big differences in image classification.   Because low level features (the output of $base(.)$) are usually common between different domains, they should be copied to the new network without any modification. Since $h_p(.)$ is part of the pre-trained network, one may need a regressor $re(.)$ on the $h_b(.)$ to get classification accuracy in $D_c$, but it is not necessary in order to minimize equation~\eqref{eq1}. Optionally, $re(h_b(base(.)))$ can be fine-tuned to get higher accuracy.

\subsubsection{Stage 2: Synthesizing Domain $D_s$} 
The second stage needs to construct the loosely labeled synthetic domain $D_s$ to approximate the distribution of the target domain $D_t$ through 
an automatic algorithm.  

\par
First, a weakly labeled method \cite{ref26} will work with hierarchic object notion \cite{ref34} to construct an initial synthetic domain $D_{init}^s$. Specifically, it needs to collect sets of images with different topics from the internet and then utilize a hierarchy notion to organize its category structure. Unfortunately, these processed images 
are coming with two problems. One is that original internet images are usually containing too many irrelevant objects to train the network directly. The other one is that true labels of those images are not matched well with image topics. For example, an image with topic Car may have a huge airplane in the middle. It is difficult 
to use this image to train a Car classifier. 

\par
So, in the second step, equation~\eqref{eq2} will be employed to keep regions that are highly related to their image topic.
\begin{align}
\begin{split}
D_{keep}^s =  &\{(x_i^p, y_i^p)|(X_i, Y_i, C_i)=\mathbb{H}(x_i^s, y_i^s),   x_i^p \in X_i, \\ & y_i^p \in Y_i, c_i^p \in C_i, c_i^p \geq \lambda,(x_i^s, y_i^s) \in D_{init}^s \}
\end{split}
\label{eq2}
\end{align}  
Where $x_i^s$ and $y_i^s$ are images and topics from $D_{init}^s$, $\mathbb{H}(.)$ is an selection function that will output a set of selected regions $X_i$, a set of topics $Y_i$ and a set of confidence values $C_i$. Each region has one topic and a confidence value. $\mathbb{H}(.)$ can be implemented by Instance Segmentation methods. $D_{keep}^s$ contains regions from $D_{init}^s$, which are highly related to their topics with confidence higher than $\lambda$. Please see Section~\ref{secIMP} for implementation details.

\par
Third, equation~\eqref{eq3} is employed to form the $D_s$ by using $D^s_{keep}$ and the hierarchical object notion. 
\begin{align}
\begin{split}
D_s = \{(x_j^{t},t)|D_{keep}^t= \mathbb{M}(D_{keep}^s,t), \\ t\in T, x_j^t \in D_{keep}^t \}
\end{split}
\label{eq3}
\end{align}
Where $\mathbb{M}(.)$ puts regions in $D^s_{keep}$ of different topics into a specific hierarchy-object-based set $D_{keep}^t$, $e.g.$, regions of topics including pedestrian, woman, man, boy, and girl will be merged into the person category (hierarchy object). $T$ is the target category set of $D_t$. $x_j^t$ is the j-th image that belongs to the t-th hierarchy object set $D_{keep}^t$. $D_s$ is the set of all $D_{keep}^t$. Please see Section~\ref{secIMP} for implementation details.

\subsubsection{Stage 3: Classifier Training} The Last stage utilizes the synthesized $D_s$ and transfer learning technologies to train a randomly initialized classifier, $h_c(.)$. So that we can approximate $\mathbb{P}^t_h$ with $\mathbb{P}^s_h$ by letting $h(.)=h_c(h_b(base(.)))$.

\par 
The trained $h_b(.)$ and the copied $base(.)$ are employed to extract features from input images. Those features are used to train a randomly initialized classifier $h_c(.)$. All images are from $D_s$ (the synthetic domain) only. They are preprocessed by specially designed methods to decrease the huge distribution variance between the synthetic domain ($D_s$) and the target domain ($D_t$) by changing pixel level attributes, $e.g.$, brightness, grayscale, contrast, and so on. Plus, $h(.)$ can be further fine-tuned on $D_s$ if necessary. The train dataset is $D_s$, and the validation and test dataset are from $D_t$. If the validation accuracy is much worse than training accuracy, then the preprocessing method should be improved because the distribution gap between $D_s$ and $D_t$ is still too big.

\par 
For the classifier training, different optimization function may lead to different generalization ability. In this work, we tested the conventional cross entropy loss \cite{ref18} and the focal loss \cite{ref38}.

\par
Equation~(\ref{eq4}) has shown the conventional cross entropy loss.
\begin{align}
L_{c} & = -\frac{1}{N} \sum_{j=1}^N (\log{{p}^{t}_j})
\label{eq4}
\end{align}
Where the ${p}^{t}_j$ is the predicted probability of the true label $t \in T$, $T$ is a set of all target categories in $D_s$, $N$ is the 
number of training images.

\par
Equation~\eqref{eq5} has shown the original focal loss \cite{ref38}.
\begin{align}
L_{f} & = -\frac{1}{N}\sum_{j=1}^N (1-{p}^{t}_j)^\gamma\log{({p}^{t}_j)}\label{eq5}
\end{align}
Where $\gamma$ is a parameter to smoothly adjust the rate in which easy examples are down weighted.

\par
After training, $base(.)$, $h_b(.)$ and $h_c(.)$ work together will be $h(.)=h_c(h_b(base(.)))$, which can be directly applied to the target domain $D_t$ with strong generalization ability. The key point is that the domain $D_t$ is used for model selection, early stop and test only.

\par
Fig.~\ref{fig2} has shown the main pipeline of CEDG algorithm which has been listed in Algorithm~\ref{alg:1} whose implementation has been described in Section~\ref{secIMP}.

\begin{algorithm}[htb]  
\scriptsize
  \renewcommand{\algorithmicrequire}{\textbf{Input:}}
  \renewcommand{\algorithmicensure}{\textbf{Output:}}
  \caption{Cost-Effective Domain Generalization}
  \label{alg:1}
  \begin{algorithmic}[1]
    \REQUIRE Bottom, middle and top parts of a pretrained network, denoted as $base(.)$, $h_p(.)$, $re(.)$ respectively,  a random initialized network $h_b(.)$ with same input and output dimensions as $h_p(.)$, a random initialized classifier $h_c(.)$ which can accept the output of $h_b(.)$ as input, a common domain $D_c$, a target domain $D_t$.
    \ENSURE $h_b(.)$, $h_c(.)$
    \STATE  \textbf{Stage 1}.Transfer prior knowledge of $h_p(.)$ to $h_b(.)$ in $D_c$ by minimizing equation~\eqref{eq1}. Optionally, one can fine tune $re(h_b(base(.)))$ on $D_c$ to get higher accuracy. 
    \STATE \textbf{Stage 2}. A three-step strategy for the construction of $D_s$. Use image spider scripts to collect loosely and weakly labeled internet images as the initial synthetic domain $D_{init}^s$. Keep meaningful pairs $(x_i^p, y_i^p)$ by equation~\eqref{eq2}. Synthesize $D_s$ through equation~\eqref{eq3}. 
	\STATE \textbf{Stage 3}. Restart this stage with better preprocessing methods if the validation accuracy on $D_t$ is not good enough.
	\WHILE{not converge}
    \FORALL{$(x_j^{t},t) \in D_s$}
    \STATE  Generate a new $\bar{x_j^{t}}$ by preprocessing $x_j^{t}$. (Please refer to Section~\ref{sec:train-hc} for detailed explanation.) 
    \STATE  Feed the $\bar{x_j^{t}}$ and $t$ to $h_c(h_b(base(.)))$, argmin equation~\eqref{eq5} $\rightarrow$ $h_c(.)$ (or/and optimize $h_b(base(.))$ if necessary).
    \STATE  Test the generalization ability of $h_c(.)$ on the validation dataset of domain $D_t$.
    \ENDFOR 
    \ENDWHILE
    \STATE \textbf{return} $base(.)$, $h_b(.)$, $h_c(.)$
  \end{algorithmic}  
\end{algorithm}

\section{An Implementation of Cost-Effective Domain Generalization}\label{secIMP}
\begin{figure}
\centering
  \includegraphics[width=0.5\textwidth]{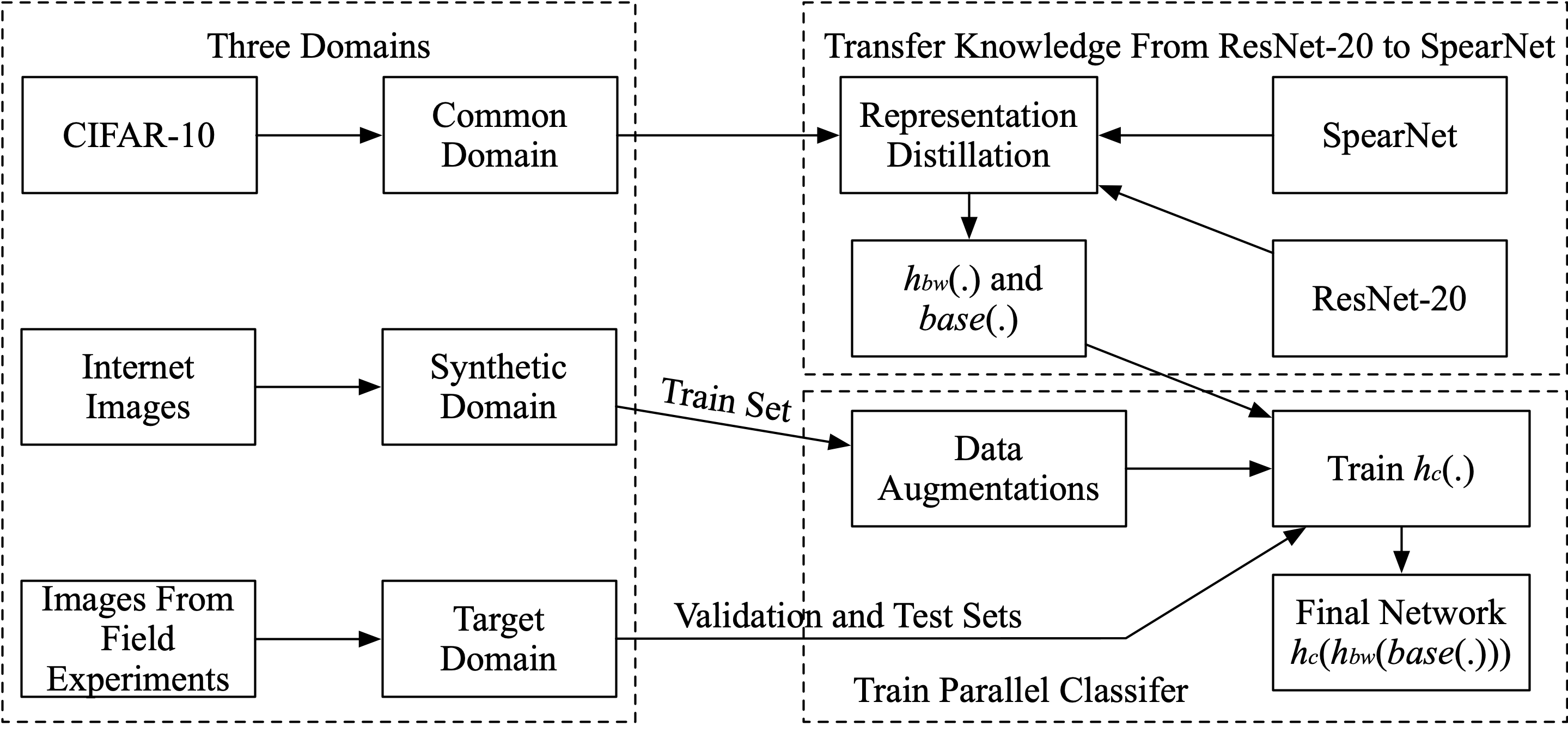} %
  \caption{ An implementation of Cost-Effective Domain Generalization.} 
  \label{fig:ImpCEDG}
\end{figure}

Fig.~\ref{fig:ImpCEDG} shows an implementation of Cost-Effective Domain Generalization (CEDG) which is explained in detail in following three subsections. All programs run on the Torch 7 framework \cite{ref37} and an NVIDIA 1080Ti GPU with 11 GB memory. The resolution of $D_c$ (Common Domain), $D_s$ (Synthetic Domain), and $D_t$ (Target Domain) are all 32 $\times$ 32 pixels.

\subsection{Three Domains}
\textbf{\slshape Common Domain $D_c$}. The Common Domain $D_c$ is CIFAR-10 dataset~\cite{cifar10} which contains 50000 images for equation~\eqref{eq1} optimization (Stage 1 of Algorithm~\ref{alg:1}), and 10000 images of test set is used for $h_b(.)$ selection. Model $re(h_b(base(.)))$ with the highest testing accuracy will be saved for Stage 3 of Algorithm~\ref{alg:1}.

\textbf{\slshape Synthetic Domain $D_s$}. This part is corresponding to Stage 2 of Algorithm~\ref{alg:1}.
The Synthetic Domain $D_s$ is generated from $D^s_{init}$. $D^s_{init}$ is a set includes 25781 loosely and weakly labeled images that are separated into 15 topics whose details have been presented in Table~\ref{tab2}. These images are collected from internet by using the 15 topics. The corresponding categories of CoCo\cite{coco} have been used to find topic-related regions by Mask-Rcnn\cite{ref17} which is $\mathbb{H}(.)$ in equation~\eqref{eq2}. Note, different instance segmentation methods can also be used to implement $\mathbb{H}(.)$. $\lambda$ in equation~\eqref{eq2} is 0.7 and the proposal region number of Mask-Rcnn is 3000. For example, regions with scores higher than 0.7 in the category of person will be saved into $D^s_{keep}$ if the input images are sampled from one of the following topics: soldier; pedestrian; boy; girl; man; woman. 

\begin{table}[htb]
\scriptsize
\centering
\caption{\newline Detailed Image Topics of $D^s_{init}$.}
\begin{tabular}{c c c c }
\hline \hline
Image topic                & CoCo                & Images      &Regions \\ \hline
Pet                        & cat, dog, bird                & 1489            &1897          \\ 
Pedestrian                 & person                        & 1606            &6009          \\ 
Mixer car                  & car                           & 1855            &2373          \\ 
Car                        & car                           & 1460            &1571          \\ 
Military truck             & truck, bus                    & 1374            &1642          \\ 
Military off-road vehicle  & car                           & 1469            &1582          \\ 
Truck                      & truck, bus                    & 1482            &1702          \\ 
Amphibious armored vehicle & airplane, train, boat         & 1444            &1570          \\ 
Wheeled armored vehicle    & airplane, train, boat         & 1405            &1404          \\ 
Goat                       & sheep                         & 1483            &1950          \\ 
Cattle                     & cow                           & 1483            &1606          \\ 
Off-road vehicle           & car                           & 2187            &3884          \\ 
Tank                       & airplane, train, boat         & 4058            &4166          \\ 
Armored personnel carrier  & airplane, train, boat         & 829             &854           \\ 
Soldier                    & person                        & 2157            &7755          \\ \hline
\end{tabular}
\label{tab2} 
\end{table}

\par
According to Table~\ref{tab3} and equation~\eqref{eq3}, images in $D^s_{keep}$ have been regrouped and labeled with their target categories. Table~\ref{tab3} has listed the size of each categories in $D_s$ and $D_t$. Images in $D_s$ have been resized into 32$\times$32 through bilinear interpolation and histogram equalized. 

\begin{table*}[htb]
\scriptsize
\centering
\caption{\newline Target Categories and Merged Topics of $D_s$.}
\begin{tabular}{c c c c c}
\hline\hline
Target Category & Merged Topic                                    & $D_s$ Size & $D_t$ Size \\ \hline
Person          & pedestrian, soldier                             & 13764     & 10770    \\ 
Wheeled Vehicle & car, military off-road vehicle, off-road vehicle, truck, military truck, mixer car   & 12754     & 10035     \\ 
Tracked Vehicle & armored personnel carrier, amphibious armored vehicle, tank, wheeled armored vehicle & 7994     & 1483     \\ 
Other           & pet, cattle, goat        & 5453     & 76      \\ \hline
\end{tabular}
\label{tab3} 
\end{table*}

\par
Since images from internet is high in resolution, we don't know where exactly the target is. Plus, we cannot 100\% sure that images are relative to their topic. So, $D^s_{init}$ is a loosely and weakly labeled dataset. Since selected by Mask-Rcnn, which is not 100\% correct in region proposals, $D^s_{keep}$ and $D_s$ are both loosely labeled datasets.   

\textbf{\slshape Target domain $D_t$}. The Target Domain $D_{t}$ is a real-world image dataset containing 22364 hand labeled images which are all sampled by WSN nodes during our field experiments. The images are all down sampled to $32 \times 32$ and have been histogram equalized. It has four categories as shown in Table~\ref{tab3}. These images are collected from different field experiments and shot by digital cameras, SLR cameras or EVIL. The whole target domain has been randomly divided into a validation set and a test set equally. The validation set has been used to select the best classifier.

\par
The standard color normalization has been applied to $D_s$ and $D_t$ by means and stds that are calculated from the histogram equalized training dataset of CIFAR-10 (the Common Domain $D_c$), which are $[136.2, 134.7, 118.9]$, $[73.9, 71.3, 76.1]$ respectively.

\subsection{Transfer Knowledge from ResNet-20 to SpearNet}
In Stage 1 of Algorithm~\ref{alg:1}, a pre-trained model ResNet-20 is used to guide the representation learning of $h_b(.)$. The pre-trained network ResNet-20 was trained on CIFAR-10 using hyperparameters listed in Table~\ref{tab1} in column Pre-train with cross entropy loss. It can reach 91.64\% testing accuracy on the Common Domain $D_c$ with approximately 41 million multiplications per prediction. It contains three groups of residual block with 16 filters, 32 filters, and 64 filters respectively. The weight optimization algorithm is the standard SGD whose hyperparameters is shown in Table~\ref{tab1}. 

\begin{table}[htb]
\caption{\newline SGD Hyperparameters at Different Stages.}
\scriptsize
\centering
\begin{tabular}{c c c c}
\hline\hline 
 & Pre-train & Stage 1 & Stage 3 \\ \hline
Batch size & 128 & 128 & 128 \\ 
Learning rate & \begin{tabular}[c]{@{}c@{}} 0$\sim$80 epochs, 0.1\\ 81$\sim$119 epochs, 0.01\\ 120$\sim$ epochs, 0.001\end{tabular} & \begin{tabular}[c]{@{}c@{}} 0$\sim$74 epochs, 0.1\\ 75$\sim$124 epochs, 0.01\\ 125$\sim$ epochs, 0.001\end{tabular} & 0.0005 \\ 
Weight decay &1e-4 & 1e-5 & 0 \\ 
Momentum & 0.9 & 0.9 & 0 \\ 
Dampening & 0 & 0 & 0 \\ 
Nesterov & true & true & false \\ \hline
\end{tabular}\\
\label{tab1} 
\end{table}

\par
\begin{figure*}[htb]
\centering
  \includegraphics[width=1\textwidth]{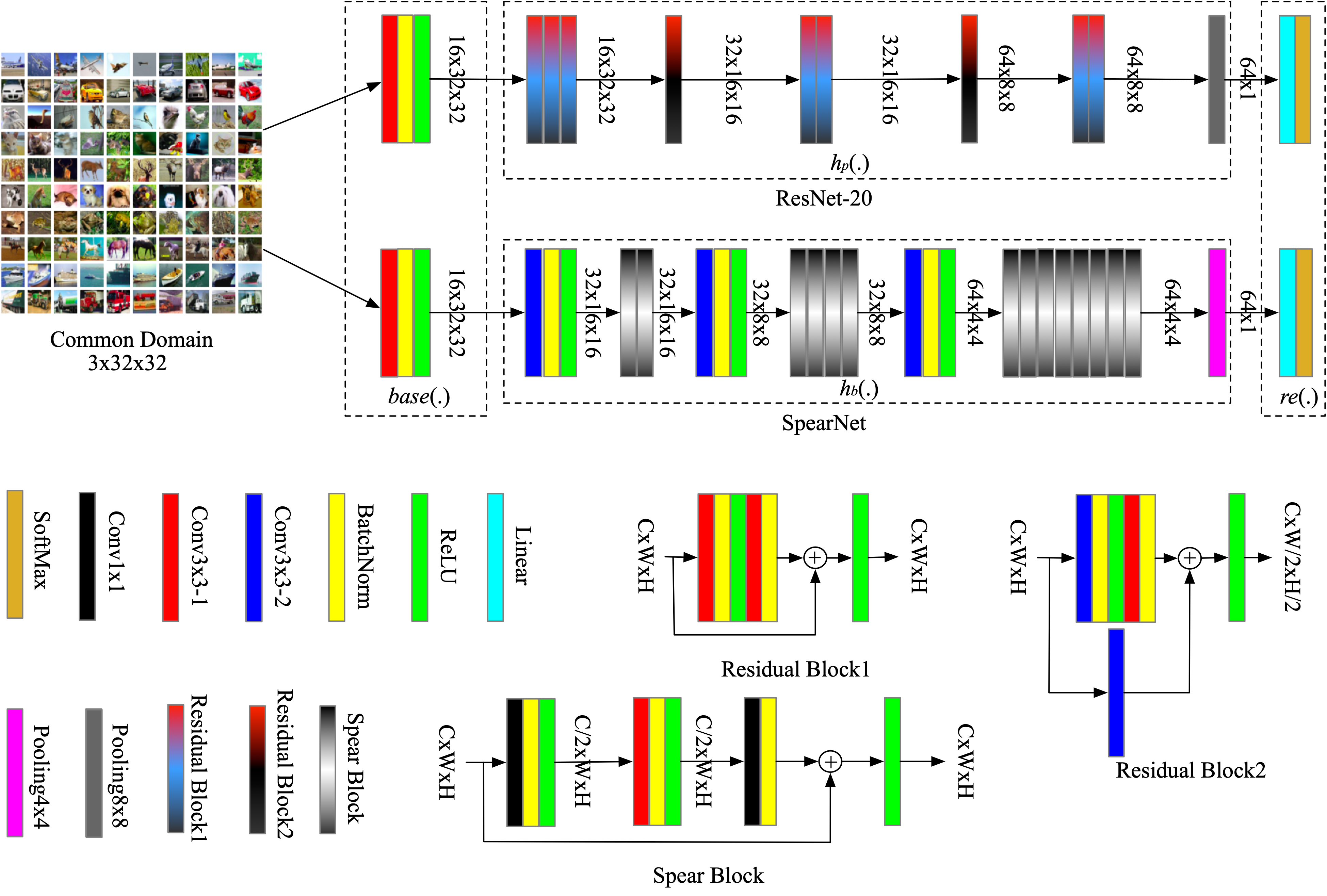} %
  \caption{ Networks used in Stage 1 of Algorithm~\ref{alg:1}.} 
  \label{fig:ResNet-20}
\end{figure*}

As shown in Fig.~\ref{fig:ResNet-20}, the network between $base(.)$ and $re(.)$ in ResNet-20, namely $h_p(.)$, has been redesigned into $h_b(.)$ of SpearNet, a kind of deep and fast network. To minimize equation~\eqref{eq1}, the pooling layer is included into $h_p(.)$ and $h_b(.)$. SpearNet uses same $base(.)$ and $re(.)$ as ResNet-20. In ResNet-20, the first convolutional layer of a residual block is used to change the channel number of features if necessary. In Fig.~\ref{fig:ResNet-20}, Conv$a$x$b$-$c$ is a convolution with kernel size $a \times b$ and stride $c$. CxWxH in Residual/Spear Block specifies the input/output dimension of the block which is C channels , W width and H height. Pooling4x4 and Pooling8x8 are using averaging layer with kernel size 4x4 and 8x8 respectively.

\par
Equation~\eqref{eq1} is optimized on the Common Domain $D_c$ which is the training dataset of CIFAR-10 \cite{cifar10} with SGD and hyperparameters listed in Table~\ref{tab1} (Stage 1). By using same hyperparameters but learning rate is 0.0001, $re(h_b(base(.)))$ is fine turned on CIFAR-10 training dataset using cross entropy loss based on the best $h_b(.)$ which has smallest equation~\eqref{eq1} testing value. Finally, the network of $h_b(.)$ without the pooling layer (noted as $h_{bw}(.)$) and $base(.)$ are used in the training of Stage 3 in Algorithm~\ref{alg:1}. 

\subsection{Train A Parallel Classifier}\label{sec:train-hc}
The classifier $h_c(.)$ includes four groups of parallel linear layers as shown in Table~\ref{tab6}. The outputs of these groups are concatenated into a vector with 4 elements. The L1 normalize layer normalizes the L1-norm of the vector into unit. We use ReLU activators after first linear layer in each group of $h_c(.)$ except the last linear layer. The output of $h_c(.)$ is further processed by SoftMax to get possibility predictions. $h_{bw}(.)$ ($h_b(.)$ without pooling layer) and $base(.)$ come from SpearNet trained in Stage 1.

\begin{table}[htb]
\scriptsize
\centering
\caption{\newline The Network Configuration of $h_c(.)$.}
\begin{tabular}{ c c c c}
\hline\hline
\begin{tabular}[c]{@{}c@{}}Linear(1024,64) \\ Linear(64,1)\end{tabular} & \begin{tabular}[c]{@{}c@{}}Linear(1024,64) \\ Linear(64,1)\end{tabular} & \begin{tabular}[c]{@{}c@{}}Linear(1024,64) \\ Linear(64,1)\end{tabular} & \begin{tabular}[c]{@{}c@{}}Linear(1024,64) \\ Linear(64,1)\end{tabular} \\ 
\multicolumn{4}{c}{L1 normalize layer}                                                                                                                                                                                                          \\ \hline
\end{tabular}
\label{tab6} 
\end{table}

Random Crop (RC), Vertical and Horizontal Flip (VHF), Graying (GI), Smooth (SH), Masking (MG) approaches have been employed as preprocessing methods to transfer domain $D_s$ to target domain $D_t$. 
\par
The Random crop method will crop a 24$\times$24 sub-area randomly and resize it to 32$\times$32.
\par
The smooth operation is designed to perform a fixed 3$\times$3 or 5$\times$5 circular-average filtering with the equal probability to approximate out-of-focus images. 
\par
The masking method is designed to cover targets by tree like masks through drawing vertical lines with random widths at random center points. Each line has the mean color of original image pixels covered by each line itself. The total width of all lines is 10. For example, the width of the first line is set as a random value between 1 and 10, denoted as $b_1$. Its center point is randomly picked from the whole image. The width of the second line is a random value between 1 and $10-b_1$. Its center point is randomly picked from all uncovered area. If the new line has overlaps with another line or fall outside the image, it will be discarded. The method just keeps drawing new lines until the width of lines in all reaches 10. 

To train the classifier, the focal loss (equation~\eqref{eq5}, where $\gamma=2$.) is used as the loss function. The SGD optimization algorithm is the Stage 3 hyperparameters of Table~\ref{tab1}. Category balancing weights are 0.1581, 0.1706, 0.2722, 0.3991 for Person, Wheeled vehicle, Tracked vehicle and Other respectively. The category balancing weights are calculated by equation~\eqref{cbw}.
\begin{equation}
CBW_i = \frac{\frac{1}{N_i}}{\sum_{j=0}^{3} \frac{1}{N_j} }  \label{cbw}
\end{equation}
Where $CBW_i$ is the weight of $i_{th}$ category of Synthetic Domain $D_s$, $N_i$ and $N_j$ are the image number of $i_{th}$ and $j_{th}$ category of $D_s$. As shown in Table~\ref{tab3}, there are 4 categories, and each has different number of images. The training has 3 steps which have been stopped based on the averaged value ($AVE$) of 4 category-level error rates on validation dataset (from $D_t$).

\begin{enumerate}[*]
\item \textbf{Step 1}. Each group of $h_c(.)$ shown in Table~\ref{tab6} is replaced by Linear(1024,10240), Linear(10240,64), and Linear(64,1) in sequence with ReLU activators between them. The enlarged $h_c(.)$ can capture features more easily. $h_{bw}(.)$ and $base(.)$ are not trained in this step. 
\item \textbf{Step 2}. $h_{bw}(.)$ and $base(.)$ are trained using the best enlarged $h_c(.)$ from \textbf{Step 1}. The enlarged  $h_c(.)$ is not trained in this step. 
\item \textbf{Step 3}. $h_c(.)$ of Table~\ref{tab6} is randomly initialized and trained. $h_{bw}(.)$ and $base(.)$ are from \textbf{Step 2} and not trained. 
\end{enumerate}

Finally, network $h_c(h_{bw}(base(.)))$ can be fine turned if necessary. The final network has the parallel classifier (Table~\ref{tab6}) as $h_c(.)$, the middle part of SpearNet ($h_b(.)$ without the pooling layer) as $h_{bw}(.)$, the low-level networks of ResNet-20 as $base(.)$. The final network is trained on Common Domain $D_c$ (CIFAR-10) and Synthetic Domain $D_s$ (loosely and weakly labeled internet images). And the final network is tested on Target Domain $D_t$ (images from field experiments).

\section{Experiments and Results}\label{secEXP}
\par
For the knowledge transfer From ResNet-20 to SpearNet (Stage 1 of Algorithm~\ref{alg:1}), the testing accuracy values before and after fine tuning have been listed in Table~\ref{tab5}, as well as the estimated numbers of multiplication that are needed by neural networks for one single prediction. Except the first one, the meanings of each column are number of required multiplications for one prediction, parameter number, testing accuracy before and after fine tuning, the compression rate in number of multiplications respectively. Although SpearNet has lower accuracy than ResNet-20, the number of required multiplications for SpearNet is only 7 million, which is much less than ResNet-20.
\par

\begin{table}[htb]
\scriptsize 
\centering
\caption{\newline Accuracy and Efficiency of Networks. MPN, PAS, COM are number of multiplications, parameters, compression rate respectively.}
\begin{tabular}{c c c c c c}
\hline\hline
Version     &MPN  &PAS    &Before     &After      &COM\\ 
\hline
ResNet-20 &41M  &273K   &\multicolumn{2}{c}{91.64\%}  &$\sim$ 1 \\ 
SpearNet    &7M   &164K   &86.28\%    &87.20\%    &$\sim$ 5.86 \\ 
\hline
\end{tabular}
\label{tab5} 
\end{table}


\par
Different kinds of $h_c(.)$ have been tested. The model shown in Table~\ref{tab6} is noted as A1. A2 and A3 are same as A1 but use L2 normalize layer and SoftMax layer respectively instead of L1 normalize layer. A4 is a modified $re(.)$ of SpearNet (Fig.~\ref{fig:ResNet-20}) which has replaced its last two layers with Linear(64,4) and L1 normalize layer. Similar to A1, the outputs of A2$\sim$4 are all processed by another SoftMax. So, A3 has two SoftMax layers. All $h_c(.)$s have been trained on $D_s$ by SGD using hyperparameters of Stage 3 listed in Table~\ref{tab1} with cross entropy loss (equation~\eqref{eq4}). Total epoch number is 200 where $h_{bw}(.)$ ($h_b(.)$ for A4) and $base(.)$ are not trained. The averaged value ($AVE$) of 4 category-level error rates on validation dataset (from $D_t$) of each epoch has been recorded as well as their error rate ($ER$) on whole validation dataset. $AVE$ metric indicates how accurate the classifier is in classifying each category. $ER$ indicates the overall accuracy of the classifier.
\par 
Equation~\eqref{SUM} has been used to find best $h_c(.)$s, where $EPN$, $EPS$ and $EPE$ are total number of epochs, the start epoch and the end epoch which are 200, 1 and 200 in this experiment. Smaller $SUM$ means $h_c(.)$ learns quicker because $h_c(.)$ tends to have smaller $AVE+ER$ at each epoch. A1$\sim$4 models have $SUM$ 0.3924, 0.4032, 0.5324 and 0.6172 respectively. So, A1 (Table~\ref{tab6}) is used in $h_c(h_{bw}(base(.)))$ in this work. 
\begin{equation}
SUM = \frac{1}{EPN} \sum_{ep=EPS}^{EPE} (AVE_{ep}+ER_{ep}) \label{SUM}
\end{equation}



\par
Different combinations of data augmentations have been tested where $h_{bw}(.)$ and $base(.)$ are from SpearNet that was trained in Stage 1; $h_c(.)$ (Table~\ref{tab6}) is randomly initialized. The train has used hyperparameters of Stage 3 listed in Table~\ref{tab1} where the whole network $h_c(h_{bw}(base(.)))$ has been iterated 3000 epochs in each test. Note, $h_{bw}(.)$ and $base(.)$ are not trained. The loss function is the focal loss (equation~\eqref{eq5}, where $\gamma=2$.). Table~\ref{tab7} lists the best validation $AVE$ values we have reached with each combination of data augmentations. In each test, images have been randomly selected from domain $D_s$ followed with histogram equalization and the color normalization. Note, category balancing weights ($CBW$) is not used. And then, they have been processed by each data augmentation in sequence with 0.5 skipping possibility. From Table~\ref{tab7}, we believe that all data augmentations together can help to transfer domain $D_s$ to domain $D_t$ and help to train a robust $D_t$ images classifier since the model has the best validation $AVE$ values. So, all data augmentations are used in the training of the final network.   
\begin{table}[htb]
\scriptsize 
\centering
\caption{\newline Data Augmentation Tests. Detailed explanations of each augmentation can be found in the Section~\ref{sec:train-hc}.}
\begin{tabular}{ c c c c c c }
\hline\hline
\begin{tabular}[c]{@{}c@{}}Random\\ Crop\end{tabular} & \begin{tabular}[c]{@{}c@{}}Vertical and \\ Horizontal Flip\end{tabular} 	& Graying & Smooth 	& Masking	& $AVE$    \\ \hline
				&								      &			&			&			&  0.1720 \\ 
	$\surd$	&								      &			&			&			& 0.1768 \\
	$\surd$	&				$\surd$			&			&			&			& 0.1876 \\ 
	$\surd$	&				$\surd$			&	$\surd$&			&			& 0.1552 \\ 
	$\surd$	&				$\surd$			&	$\surd$&	$\surd$&			& 0.1554 \\ 
	$\surd$	&				$\surd$			&	$\surd$&	$\surd$&	$\surd$& \textbf{0.1405} \\ \hline
\end{tabular}
\label{tab7} 
\end{table}
\par

\begin{figure}
\scriptsize 
\centering
\includegraphics[]{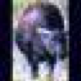}
\includegraphics[]{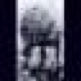}
\includegraphics[]{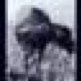}
\includegraphics[]{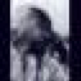}
\caption{Misclassified Other images in the test dataset which have been histogram equalized.}
\label{fig-Other}
\end{figure}

Table~\ref{tab:whole_train} shows the accuracies of the training of the final network as well as the confusion matrices have been shown in Table~\ref{tab-confusion}. The best model of each step has been selected based on the validation $AVE$. $FL$ is the averaged value of focal loss on train set. We use $AVE$ to select the best model because of the unbalanced dataset $D_t$. If we use $ER$ instead of $AVE$, then the best classifier may has very low accuracy in the Other category. Since the category balancing weights weight the Other category most, the global correct rate has been decreased. Because the Other category contains too much hard samples relative to its total amount (Fig.~\ref{fig-Other} has shown misclassified test images in the Other category), the classifier must sacrifice global correct rate to get a higher Other-category accuracy. But the employment of the category balancing weights is necessary. Otherwise, the classifier will simply skip the Other category and focus on the other three categories to get higher global accuracy.  

\begin{table}[htb]
\centering
\caption{\newline Classification accuracies of each steps. FL, AVE, ER are the averaged focal loss on train set, average of category-level error rates, error rate respectively. All values are averaged on corresponding dataset. }
\begin{tabular}{cccccc}
\hline\hline
\multirow{2}{*}{Step} & Train  & \multicolumn{2}{c}{Validation} & \multicolumn{2}{c}{Test} \\ 
                      & $FL$     & $AVE$            & $ER$            & $AVE$         & $ER$         \\ \hline
1                     & 0.4604 & 0.1526         & \textbf{0.1600}        & 0.1575      & \textbf{0.1596}     \\
2                     & 0.3990 & \textbf{0.1229}         & 0.1851        & 0.1348      & 0.1908     \\
3                     & \textbf{0.3916} & 0.1293         & 0.1849        & \textbf{0.1330}      & 0.1892    \\ \hline
\end{tabular}
\label{tab:whole_train} 
\end{table}

\begin{table}[htb]
\scriptsize 
\centering
\caption{\newline Confusion Matrix.}
\begin{tabular}{ c c c c c c }
\hline\hline
$D_t$                       & Person & Other & Track & Wheel & Accuracy \\ \hline
\multirow{4}{*}{Validation} & 4690   & 281   & 43    & 371   & 87.09\%  \\ 
                            & 1      & 36    & 1     & 1     & 92.31\%  \\  
                            & 4      & 2     & 709   & 27    & 95.55\%  \\ 
                            & 155    & 127   & 1055   & 3679  & 73.36\%  \\ \hline
\multirow{4}{*}{Test}       & 4748   & 289   & 39    & 309   & 88.17\%  \\  
                            & 3      & 33    & 0     & 1     & 89.19\%  \\ 
                            & 6      & 1     & 696   & 38    & 93.93\%  \\ 
                            & 157    & 150   & 1203   & 3509  & 69.91\%  \\ \hline
\end{tabular}
\label{tab-confusion}
\end{table}

\par
Another model has validation and test $AVE$ 0.1442 and 0.1472. It has been trained with same configuration as Section~\ref{sec:train-hc} but using the cross-entropy loss (equation~\eqref{eq4}) as the loss function. It indicates that the focal loss works better than the cross-entropy loss in this application.
\par
To verify the proposed method (CEDG) do help in training efficient deep convolutional neural networks, we trained a randomly initialized $h_c(h_{bw}(base(.)))$ which has same network structure as Table~\ref{tab6} and SpearNet in Fig.~\ref{fig:ResNet-20} without the pooling layer. The configurations are:
1) using $D_s$ as training dataset with the category balancing weights, $D_t$ as validation and test dataset;
2) using all data augmentation methods together to preprocess training images;
3) using the focal loss and the SGD optimization method.
The network was first trained with Stage 1 hyperparameters listed in Table~\ref{tab1} until its convergence. And then based on the best model, we restarted the train with Stage 3 hyperparameters. As the result, the validation and test $AVE$ of the best model are 0.2009 and 0.2259. The global correct rates are 76.75\% and 75.25\% respectively which are lower than the model trained by the proposed CEDG method.

\section{Conclusion}\label{secCONCLU}
In this paper, a Cost-Effective Domain Generalization algorithm (CEDG) has been proposed to train a small-scale and efficient network for wireless sensor network (WSN) applications using loosely and weakly labeled images. The method is cost effective since it automatically collects data from the internet to construct a synthetic domain for the network training. The method is robust since it builds efficient networks out of a large pretrained network on a common domain such as CIFAR-10 so that the small networks have knowledge passed from the state-of-art pretrained network. The method guarantees the generalization ability of the trained networks since it leaves the target domain alone for validation and test only, which consists of images from field experiments. The method is WSN orientated since it is designed to classify images that are captured from WSN field experiments with limited computation budget.    
  
\par
From the experiments results, we have following conclusions, where $AVE$ is the average of category-level error rates which is important for unbalanced test data set, $SUM$ is the averaged $AVE+ER$ of each epoch which will be small if the model learns quickly. 
\begin{enumerate}[1)] 
\item Using CEDG algorithm can decrease test $AVE$ by 41.12\%, 0.1330 (with CEDG) vs. 0.2259 (without CEDG). 
\item Data augmentation methods can decrease validation $AVE$ by 28.31\%.
\item Using focal loss instead of cross entropy loss can decrease test $AVE$ by 9.65\%.
\item Adding L1 normalization layer before SoftMax layer can decrease validation $SUM$ by 2.68\% (than L2 normalize and SoftMax layers) or 26.30\% (than 2 SoftMax layers). 
\item Using parallel linear layers and removing pooling layer can decrease validation $SUM$ by 36.42\%.
\end{enumerate}

\par
By using $h_{bw}(base(.))$ of SpearNet (Fig.~\ref{fig:ResNet-20}) and $h_c(.)$ of Table~\ref{tab6}, the trained network requires much less multiplications than ResNet-20, about 7M per prediction. We are currently transplanting the network to our WSN nodes which have a Digital Signal Processor (DSP) chip. The DSP chip can do at least 600M multiplications per second which means the node is fully capable in running the network in real time. Target localization is not evaluated in this paper because the target can be detected by background subtraction in our field experiments. In all, we believe the proposed method is essential to train small and efficient deep neural networks to run on embedded systems locally with limited hand-labeled data. 

\section*{Acknowledgment}
This work is financially supported by the National Natural Science Foundation of China (61802347), the Microsystems Technology Key Laboratory Foundation of China (61428040104, 6142804010203), the Zhejiang Province welfare technology applied research project (LGF20H180002, LY18F020031), the Compound Sensor Technology Foundation (6141A011116XX), and in part by Science and Technology on Microsystem Laboratory, Shanghai Institute of Microsystem and Information Technology, Chinese Academy of Sciences. We would like to thank Jianing Zhu for her help on image spider script and Qiliang Zhan for wild images annotation.

\section*{References}
\bibliographystyle{elsarticle-num}
\bibliography{references}

\begin{thebibliography}{10}
\expandafter\ifx\csname url\endcsname\relax
  \def\url#1{\texttt{#1}}\fi
\expandafter\ifx\csname urlprefix\endcsname\relax\def\urlprefix{URL }\fi
\expandafter\ifx\csname href\endcsname\relax
  \def\href#1#2{#2} \def\path#1{#1}\fi

\bibitem{ws3}
Q.~Zhou, B.~Li, Z.~Kuang, D.~Xie, G.~Tong, L.~Hu, X.~Yuan, A quarter-car
  vehicle model based feature for wheeled and tracked vehicles classification,
  Journal of Sound and Vibration 332~(26) (2013) 7279--7289.

\bibitem{ws8}
Y.~K. An, S.-M. Yoo, C.~An, B.~E. Wells, Rule-based multiple-target tracking in
  acoustic wireless sensor networks, Computer Communications 51 (2014) 81--94.

\bibitem{ws5}
Q.~Zhou, G.~Tong, B.~Li, X.~Yuan, A practicable method for ferromagnetic object
  moving direction identification, IEEE Transactions on Magnetics 48~(8) (2012)
  2340--2345.

\bibitem{ws9}
J.~Lan, S.~Nahavandi, T.~Lan, Y.~Yin, Recognition of moving ground targets by
  measuring and processing seismic signal, Measurement 37~(2) (2005) 189--199.

\bibitem{ws2}
Q.~Zhou, B.~Li, H.~Liu, S.~Chen, J.~Huang, Microphone-based vibration sensor
  for ugs applications, IEEE Transactions on Industrial Electronics 64~(8)
  (2017) 6565--6572.

\bibitem{ws6}
P.~E. William, M.~W. Hoffman, Classification of military ground vehicles using
  time domain harmonics' amplitudes, IEEE Transactions on Instrumentation and
  Measurement 60~(11) (2011) 3720--3731.

\bibitem{ws4}
Q.~Zhou, G.~Tong, D.~Xie, B.~Li, X.~Yuan, A seismic-based feature extraction
  algorithm for robust ground target classification, IEEE Signal Processing
  Letters 19~(10) (2012) 639--642.

\bibitem{ws1}
H.~Liu, B.~Li, X.~Yuan, Q.~Zhou, J.~Huang, A robust real time
  direction-of-arrival estimation method for sequential movement events of
  vehicles, Sensors 18~(4) (2018) 992.

\bibitem{ws10}
L.-C. Chen, M.~D. Collins, Y.~Zhu, G.~Papandreou, B.~Zoph, F.~Schroff, H.~Adam,
  J.~Shlens, Searching for efficient multi-scale architectures for dense image
  prediction, arXiv preprint arXiv:1809.04184.

\bibitem{zeng2019improved}
N.~Zeng, Z.~Wang, H.~Zhang, K.-E. Kim, Y.~Li, X.~Liu, An improved particle
  filter with a novel hybrid proposal distribution for quantitative analysis of
  gold immunochromatographic strips, IEEE Transactions on Nanotechnology 18
  (2019) 819--829.

\bibitem{zeng2018facial}
N.~Zeng, H.~Zhang, B.~Song, W.~Liu, Y.~Li, A.~M. Dobaie, Facial expression
  recognition via learning deep sparse autoencoders, Neurocomputing 273 (2018)
  643--649.

\bibitem{zeng2020deep}
N.~Zeng, H.~Li, Z.~Wang, W.~Liu, S.~Liu, F.~E. Alsaadi, X.~Liu,
  Deep-reinforcement-learning-based images segmentation for quantitative
  analysis of gold immunochromatographic strip, Neurocomputing.

\bibitem{zhou2020residual}
Q.~Zhou, P.~Tao, X.~Li, S.~Chen, F.~Zhang, H.~Hu, Residual-recursion
  autoencoder for shape illustration images, arXiv preprint arXiv:2002.02063.

\bibitem{hu2019mc}
H.~Hu, Y.~Zheng, Q.~Zhou, J.~Xiao, S.~Chen, Q.~Guan, Mc-unet: Multi-scale
  convolution unet for bladder cancer cell segmentation in phase-contrast
  microscopy images, in: 2019 IEEE International Conference on Bioinformatics
  and Biomedicine (BIBM), IEEE, 2019, pp. 1197--1199.

\bibitem{hu2019background}
H.~Hu, C.~Du, Q.~Guan, Q.~Zhou, P.~Vera, S.~Ruan, A background-based data
  enhancement method for lymphoma segmentation in 3d pet images, in: 2019 IEEE
  International Conference on Bioinformatics and Biomedicine (BIBM), IEEE,
  2019, pp. 1194--1196.

\bibitem{zhou2019towards}
Q.~Zhou, C.~Zhou, H.~Hu, Y.~Chen, S.~Chen, X.~Li, Towards the automation of
  deep image prior, arXiv preprint arXiv:1911.07185.

\bibitem{hu2018fast}
H.~Hu, C.~Luo, Q.~Guan, X.~Li, S.~Chen, Q.~Zhou, A fast online multivariable
  identification method for greenhouse environment control problems,
  Neurocomputing 312 (2018) 63--73.

\bibitem{kong2019classification}
Y.~Kong, J.~Gao, Y.~Xu, Y.~Pan, J.~Wang, J.~Liu, Classification of autism
  spectrum disorder by combining brain connectivity and deep neural network
  classifier, Neurocomputing 324 (2019) 63--68.

\bibitem{an2019generalization}
Z.~An, S.~Li, J.~Wang, Y.~Xin, K.~Xu, Generalization of deep neural network for
  bearing fault diagnosis under different working conditions using multiple
  kernel method, Neurocomputing 352 (2019) 42--53.

\bibitem{li2019constrained}
G.~Li, Z.~Liu, R.~Shi, W.~Wei, Constrained fixation point based segmentation
  via deep neural network, Neurocomputing 368 (2019) 180--187.

\bibitem{tii1}
L.~Li, K.~Ota, M.~Dong, Deep learning for smart industry: Efficient manufacture
  inspection system with fog computing, IEEE Transactions on Industrial
  Informatics 14~(10) (2018) 4665--4673.

\bibitem{tii2}
L.~Lyu, J.~C. Bezdek, X.~He, J.~Jin, Fog-embedded deep learning for the
  internet of things, IEEE Transactions on Industrial Informatics 15~(7) (2019)
  4206--4215.

\bibitem{tii3}
T.~Wang, L.~Qiu, A.~K. Sangaiah, G.~Xu, A.~Liu, Energy-efficient and
  trustworthy data collection protocol based on mobile fog computing in
  internet of things, IEEE Transactions on Industrial Informatics 16~(5) (2019)
  3531--3539.

\bibitem{tmc1}
J.~Yue, M.~Xiao, Coding for distributed fog computing in internet of mobile
  things, IEEE Transactions on Mobile Computing.

\bibitem{tmc2}
D.~Kim, H.~Lee, H.~Song, N.~Choi, Y.~Yi, Economics of fog computing: Interplay
  among infrastructure and service providers, users, and edge resource owners,
  IEEE Transactions on Mobile Computing.

\bibitem{nc1}
D.~Gutierrez-Galan, J.~P. Dominguez-Morales, E.~Cerezuela-Escudero,
  A.~Rios-Navarro, R.~Tapiador-Morales, M.~Rivas-Perez, M.~Dominguez-Morales,
  A.~Jimenez-Fernandez, A.~Linares-Barranco, Embedded neural network for
  real-time animal behavior classification, Neurocomputing 272 (2018) 17--26.

\bibitem{nc2}
W.~Zhang, K.~Liu, W.~Zhang, Y.~Zhang, J.~Gu, Deep neural networks for wireless
  localization in indoor and outdoor environments, Neurocomputing 194 (2016)
  279--287.

\bibitem{comp1}
S.~Han, J.~Pool, J.~Tran, W.~Dally, Learning both weights and connections for
  efficient neural network, in: Advances in neural information processing
  systems, 2015, pp. 1135--1143.

\bibitem{comp2}
S.~Han, H.~Mao, W.~J. Dally, Deep compression: Compressing deep neural networks
  with pruning, trained quantization and huffman coding, arXiv preprint
  arXiv:1510.00149.

\bibitem{ref2}
Y.~Cheng, D.~Wang, P.~Zhou, T.~Zhang, Model compression and acceleration for
  deep neural networks: The principles, progress, and challenges, IEEE Signal
  Processing Magazine 35~(1) (2018) 126--136.

\bibitem{ref4}
J.~Wu, C.~Leng, Y.~Wang, Q.~Hu, J.~Cheng, Quantized convolutional neural
  networks for mobile devices, in: Proceedings of the IEEE Conference on
  Computer Vision and Pattern Recognition, 2016, pp. 4820--4828.

\bibitem{ref32}
X.~Zhang, R.~Xiong, W.~Lin, J.~Zhang, S.~Wang, S.~Ma, W.~Gao, Low-rank-based
  nonlocal adaptive loop filter for high-efficiency video compression, IEEE
  Transactions on Circuits and Systems for Video Technology 27~(10) (2017)
  2177--2188.

\bibitem{ref12}
S.~Zhai, Y.~Cheng, Z.~M. Zhang, W.~Lu, Doubly convolutional neural networks,
  in: Advances in neural information processing systems, 2016, pp. 1082--1090.

\bibitem{ref14}
P.~Luo, Z.~Zhu, Z.~Liu, X.~Wang, X.~Tang, et~al., Face model compression by
  distilling knowledge from neurons, in: AAAI, 2016, pp. 3560--3566.

\bibitem{md}
C.~Buciluǎ, R.~Caruana, A.~Niculescu-Mizil, Model compression, in: Proceedings
  of the 12th ACM SIGKDD international conference on Knowledge discovery and
  data mining, ACM, 2006, pp. 535--541.

\bibitem{ref15}
A.~Romero, N.~Ballas, S.~E. Kahou, A.~Chassang, C.~Gatta, Y.~Bengio, Fitnets:
  Hints for thin deep nets, arXiv preprint arXiv:1412.6550.

\bibitem{ref31}
C.~Sun, A.~Shrivastava, S.~Singh, A.~Gupta, Revisiting unreasonable
  effectiveness of data in deep learning era, in: Computer Vision (ICCV), 2017
  IEEE International Conference on, IEEE, 2017, pp. 843--852.

\bibitem{ref26}
W.~Xu, J.~Wu, S.~Ding, L.~Lian, H.~Chao, Enhancing face recognition from
  massive weakly labeled data of new domains, Neural Processing Letters (2018)
  1--12.

\bibitem{ref27}
X.~Han, J.~Lu, C.~Zhao, S.~You, H.~Li, Semi-supervised and weakly-supervised
  road detection based on generative adversarial networks, IEEE Signal
  Processing Letters.

\bibitem{cifar10}
A.~Krizhevsky, G.~Hinton, Learning multiple layers of features from tiny
  images, Tech. rep., Citeseer (2009).

\bibitem{ref34}
A.~Zweig, D.~Weinshall, Exploiting object hierarchy: Combining models from
  different category levels, IEEE, 2007.

\bibitem{ref18}
A.~Krizhevsky, I.~Sutskever, G.~E. Hinton, Imagenet classification with deep
  convolutional neural networks, in: Advances in neural information processing
  systems, 2012, pp. 1097--1105.

\bibitem{ref38}
T.-Y. Lin, P.~Goyal, R.~Girshick, K.~He, P.~Doll{\'a}r, Focal loss for dense
  object detection, in: Proceedings of the IEEE international conference on
  computer vision, 2017, pp. 2980--2988.

\bibitem{ref37}
R.~Collobert, K.~Kavukcuoglu, C.~Farabet, Torch7: A matlab-like environment for
  machine learning, in: BigLearn, NIPS workshop, no. EPFL-CONF-192376, 2011.

\bibitem{coco}
T.-Y. Lin, M.~Maire, S.~Belongie, J.~Hays, P.~Perona, D.~Ramanan,
  P.~Doll{\'a}r, C.~L. Zitnick, Microsoft coco: Common objects in context, in:
  European conference on computer vision, Springer, 2014, pp. 740--755.

\bibitem{ref17}
K.~He, G.~Gkioxari, P.~Doll{\'a}r, R.~Girshick, Mask r-cnn, in: Computer Vision
  (ICCV), 2017 IEEE International Conference on, IEEE, 2017, pp. 2980--2988.

\end{thebibliography}

\parpic{\includegraphics[width=1in,clip,keepaspectratio]{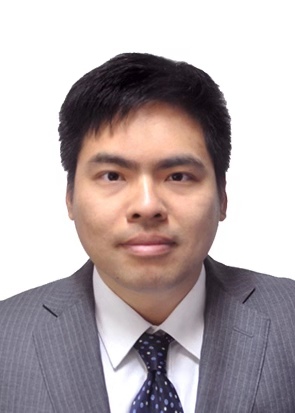}}
\noindent {\bf Qianwei Zhou} received the Ph.D. degree in communication and information 
systems from the Shanghai Institute of Microsystem and Information Technology, University of Chinese Academic and Sciences, Shanghai, China, in 2014. 
\newline \indent \indent 
In July 2014, he joined Zhejiang University of Technology, Hangzhou, China, where he is currently a Research Scientist in the College of Computer Science. And he was also a visiting scholar in Imaging Research Division, Department of Radiology, University of Pittsburgh, USA, during 08/2018-08/2019. He has published scientific papers in international journals and conference proceedings, including the IEEE TRANSACTIONS ON INDUSTRIAL ELECTRONICS, IEEE SIGNAL PROCESSING LETTERS and IEEE TRANSACTIONS ON MAGNETICS. He has been awarded a grant from the National Natural Science Foundation of China (61802347). His research interests include the crossing field of machine learning and computer-aided design, Internet-of-Thing related signal processing, pattern recognition and medical image understanding. 

\parpic{\includegraphics[width=1in,clip,keepaspectratio]{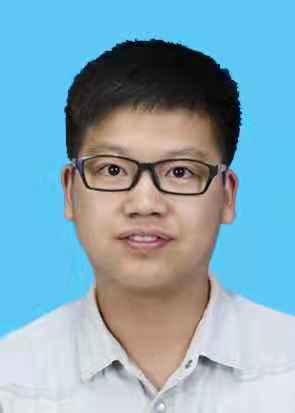}}
\noindent {\bf Yuhang Chen} received the B.S. degree in Zhejiang University of Technology, Zhejiang, China, in 2017. Now he is working towards the M.S. degree in computer science at Zhejiang University of Technology, Zhejiang, China. His major research interests are machine learning (deep learning), reinforcement learning, and transform learning. 

\parpic{\includegraphics[width=1in,clip,keepaspectratio]{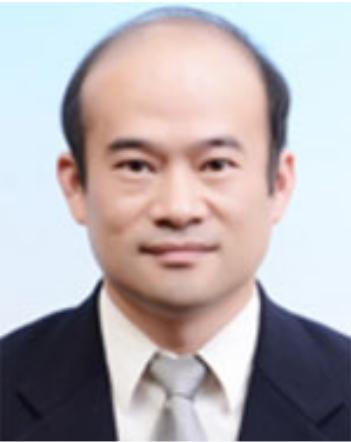}}
\noindent {\bf Baoqing Li} received the Ph.D. degree in MEMS systems from the State Key Laboratory of Transducer Technology, Shanghai Institute of Metallurgy, Chinese Academy of Sciences, Shanghai, China, in 1999. \newline \indent \indent 
In March 2006, he joined the Shanghai Institute of Microsystem and Information Technology, Chinese Academy of Sciences, Shanghai, China, where he is currently a Professor and a Tutor of Ph.D. students in the area of wireless sensor networks. From 2001 to 2005, he was with the Center of Microelectronics, New Jersey Institute of Technology, Newark, NJ, USA, working with a focus on applications of MEMS technology. 

\parpic{\includegraphics[width=1in,clip,keepaspectratio]{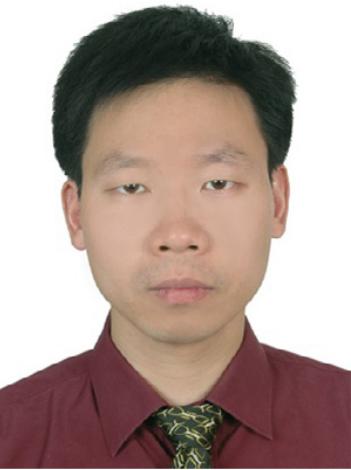}}
\noindent {\bf Xiaoxin Li} received the Ph.D. degree in computer application technology from South China University of Technology, Guangzhou, China, in 2009. 
\newline \indent \indent 
Since then, he has been a postdoctoral researcher in the Department of Mathematics, Faculty of Mathematics and Computing, Sun Yat-Sen University, Guangzhou, China. He joined Zhejiang University of Technology, Hangzhou, China, in 2013. His current research interests include deep learning, pattern recognition, and image analysis.

\parpic{\includegraphics[width=1in,clip,keepaspectratio]{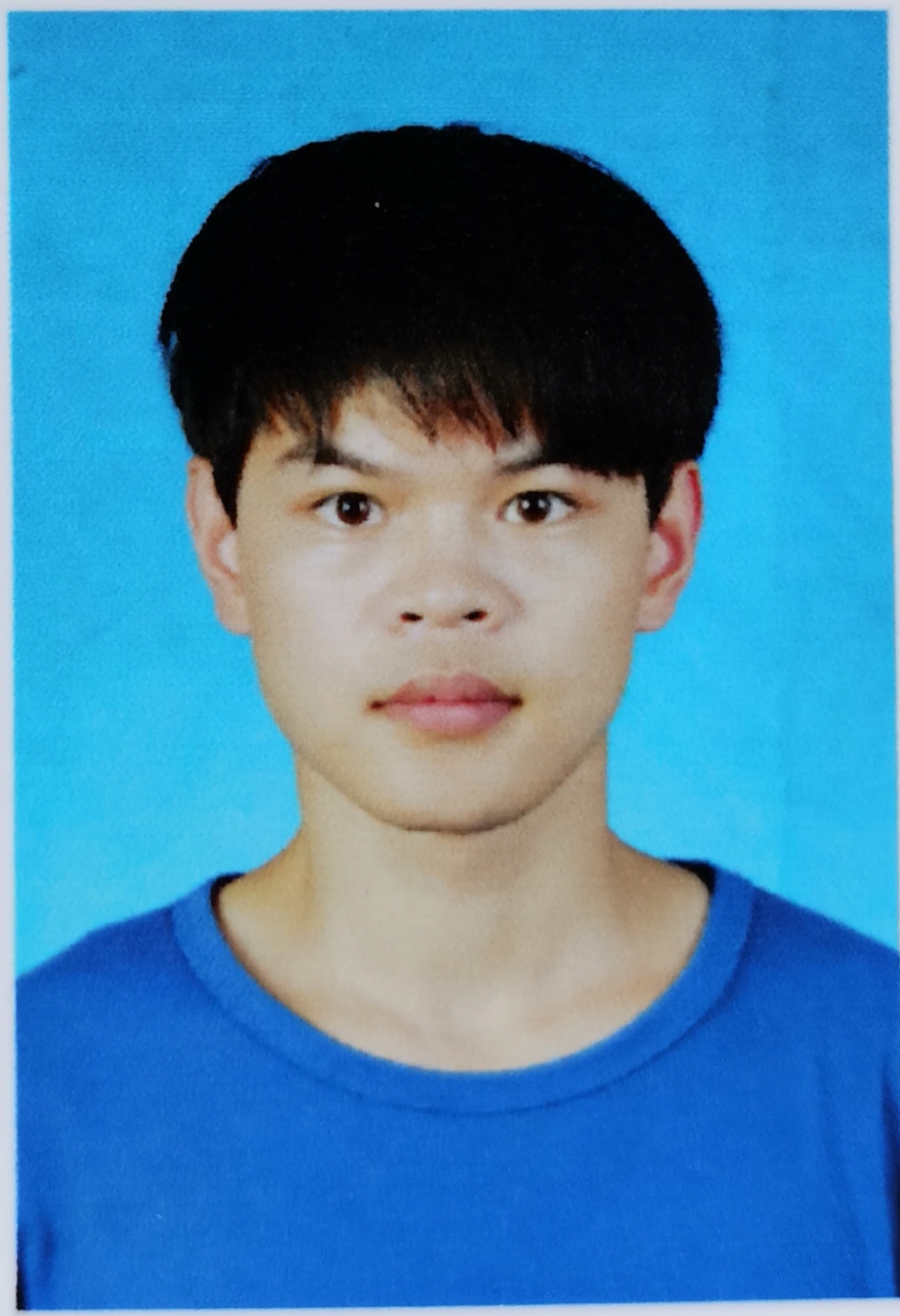}}
\noindent {\bf Chen Zhou} received the B.S. degree in LinYi University, Shandong, China, in 2018. Now he is working towards the M.S. degree in computer science at Zhejiang University of Technology, Zhejiang, China. His major research interests are machine learning (deep learning), reinforcement learning, and transform learning.

\parpic{\includegraphics[width=1in,clip,keepaspectratio]{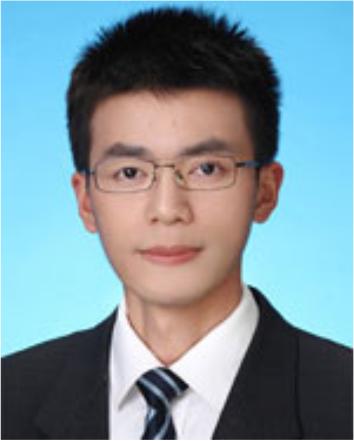}}
\noindent {\bf Jingchang Huang} received the Ph.D. degree in communication and information systems from University of Chinese Academic and Sciences, China in 2015. \newline \indent \indent 
Now he is a research fellow at Shanghai Institute of Microsystem And Information Technology, Chinese Academy of Sciences He mainly focuses on the cognitive signal processing, including but not limited to acoustic, image, air quality data, all of which comes from Internet-of-Thing related applications. His current research interests also include patterns recognition, big data and wireless sensor networks. 
\newline \indent \indent 
Dr. Huang has designed and implemented many efficient target detection and classification algorithms for unattended ground sensors system and published them in IEEE transactions and journals.

\parpic{\includegraphics[width=1in,clip,keepaspectratio]{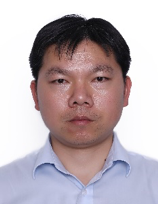}}
\noindent {\bf Haigen Hu} received the Ph.D. degree in Control theory and Control Engineering from Tongji University, Shanghai, China in 2013. \newline \indent \indent 
He is currently an associate professor in the College of Computer Science and Technology at Zhejiang University of Technology, China. And he was also a postdoctoral researcher in LITIS Laboratory, Université de Rouen, France, during 06/2018-06/2019. He has been awarded a grant from the National Natural Science Foundation of China(Grant No. 61374094) in 2013 and a grant from Natural Science Foundation of Zhejiang Province (Grant No. Y18F030084) in 2017.  His current research interests include Machine Learning (Deep learning), Neural networks, Multi-objective Evolutionary Algorithms (MOEAs), Greenhouse Environmental Control, bioinformatics, and its Applications.

\end{document}